\documentclass[letterpaper, 10 pt, conference]{ieeeconf}  % Comment this line out if you need a4paper

\IEEEoverridecommandlockouts                              % This command is only needed if 
\usepackage{subcaption}
\usepackage{graphicx}
\usepackage{mathtools}
\usepackage{color,soul} % Probably not needed when finished editing
\usepackage{textcomp} % Loading this before gensymb clears some warnings that pop up due to an error in the order gensymb loads other packages
\usepackage{gensymb}
\usepackage{bm}
\usepackage{multirow}
\usepackage{caption}
\usepackage{booktabs}
\usepackage{dblfloatfix}

\begin{document}

% \fancyhdr{\copyright 2022 IEEE.  Personal use of this material is permitted.  Permission from IEEE must be obtained for all other uses, in any current or future media, including reprinting/republishing this material for advertising or promotional purposes, creating new collective works, for resale or redistribution to servers or lists, or reuse of any copyrighted component of this work in other works.}

% \title{\LARGE \bf
% Visual Servoing for Pose Control of Soft Continuum Arm in a Structured Environment
% }

\title{Visual Servoing for Pose Control of Soft Continuum Arm in a Structured Environment}

\author{Shivani Kamtikar$^{1}$, Samhita Marri$^{2}$, Benjamin Walt$^{3}$, Naveen Kumar Uppalapati$^{4}$,\\ Girish Krishnan$^{3}$, Girish Chowdhary$^{1}$ $^{4}$  % <-this % stops a space
% \thanks{*This work was supported by NSF USDA, AIFARMS, COALESE}% <-this % stops a space
\thanks{\copyright 2022 IEEE. Personal use of this material is permitted. Permission from IEEE must be obtained for all other uses, in any current or future media, including reprinting/republishing this material for advertising or promotional purposes, creating new collective works, for resale or redistribution to servers or lists, or reuse of any copyrighted component of this work in other works. DOI: 10.1109/LRA.2022.3155821}
\thanks{This work is funded in part by AIFARMS National AI institute in agriculture supported by Agriculture and Food Research Initiative (AFRI) grant no. 2020-67021-32799/project accession no.1024178 from the USDA National Institute of Food and Agriculture, by USDA-NSF NRI grant USDA 2019-67021-28989, NSF 1830343, and by joint NSF-USDA COALESCE grant, USDA 2021-67021-34418.}
\thanks{$^{1}$Computer Science, $^{2}$ Electrical and Computer Engineering, $^{3}$ Mechanical Science and Engineering, $^{4}$ Coordinated Science Laboratory, 
University of Illinois at Urbana Champaign, USA.
        {\tt(skk7, marri2, walt, uppalap2,
        gkrishna,girishc)@illinois.edu}}%
}

% \markboth{IEEE Robotics and Automation Letters. Preprint Version. Accepted Month, Year}
% \thispagestyle{myheadings}

% \markboth{copyright}

\maketitle

% 
% \pagestyle{empty}

%%%%%%%%%%%%%%%%%%%%%%%%%%%%%%%%%%%%%%%%%%%%%%%%%%%%%%%%%%%%%%%%%%%%%%%%%%%%%%%%
\begin{abstract}
 %Despite the emergence of several soft continuum pneumatic arms, precise reaching of a target point in 3D space remains an open challenge. This is largely due to both inherent actuation  and external disturbances on the soft system. In this letter, we  present a deep neural network-based method to perform precise and smooth 3D positioning tasks on a soft arm by visual servoing. A convolutional neural network is trained to predict the actuations required to achieve a desired pose in a structured environment. Integrated and modular approaches for estimating the actuations from the image are proposed. Proportional control law is used to reduce the error between the desired and current image as seen by the camera mounted on the distal end of the soft arm. The control law makes the described approach robust to the challenges pertained with controlling a soft continuum arm. Through experiments performed on the soft arm, we compare the performance of both the methods in reaching the desired pose. Our approach is shown to be capable to reach new target objects in the structured environment. The integrated approach is robust to lighting changes, uniform loads and diminution of the soft arm. Finally, we showcase the advantage of the self supervised training method and its ease of transfer to a new environment.
%Alternate abstract
For soft continuum arms, visual servoing is a popular control strategy that relies on visual feedback to close the control loop. However, robust visual servoing is challenging as it requires reliable feature extraction from the image, accurate control models and sensors to perceive the shape of the arm, both of which can be hard to implement in a soft robot. This letter circumvents these challenges by presenting a deep neural network-based method to perform smooth and robust 3D positioning tasks on a soft arm by visual servoing using a camera mounted at the distal end of the arm. A convolutional neural network is trained to predict the actuations required to achieve the desired pose in a structured environment. Integrated and modular approaches for estimating the actuations from the image are proposed and are experimentally compared. A proportional control law is implemented to reduce the error between the desired and current image as seen by the camera. The model together with the proportional feedback control makes the described approach robust to several variations such as new targets, lighting, loads, and diminution of the soft arm. Furthermore, the model lends itself to be transferred to a new environment with minimal effort.
\end{abstract}

%%%%%%%%%%%%%%%%%%%%%%%%%%%%%%%%%%%%%%%%%%%%%%%%%%%%%%%%%%%%%%%%%%%%%%%%%%%%%%%%
\section{INTRODUCTION}
\subsection{Motivation}
Soft continuum arms (SCA) \cite{hughes2016soft} have received growing attention due to their superiority in dexterous manipulation and safe interaction with the environment. Their inherent flexibility with high degrees of freedom endows soft robots with good adaptability but raises challenges for accurate position control. The challenges in SCA control can be attributed mainly to the difficulties in modeling and sensing \cite{rus2015design} its deformed shape. Current modeling methods are either simplistic with a constant curvature assumption that work in 2D plane or valid for SCAs with short lengths \cite{george2020first}. On the other hand, Cosserat models \cite{gazzola2018forward} require expert knowledge for their implementation and therefore have been less explored by the community. In addition, even with effective models, there aren't cost-effective sensors \cite{shih2020electronic,thuruthel2019soft} to get the spatial position feedback of SCAs. 

Recent advances in visual servoing and deep learning in robots can be effectively used to overcome the limitations in both sensing and modeling of SCA. With a camera (eye-in-hand configuration) at the distal tip of the SCA acting as a feedback sensor, the pose errors can be reduced. Visual servoing using Neural Networks (NN) in conventional robotic arms has been well studied but not extensively validated on SCA because of its complex behavior. This letter proposes the use of NN for visual servoing in SCA using two approaches: \textit{integrated} and \textit{modular}. \vspace{-0.1cm}
\subsection{Related work}
Visual servoing by its name is to control a system using vision.  Classical visual servoing extracted features like points or lines using early computer vision techniques, and control was designed based on these features as seen in  \cite{espiau1992newvs}, \cite{chaumette2000solutionvs}. This limited the types of objects that can be used, the environment lighting conditions, and are heavily dependent on the reliability of feature extraction methods. The introduction of using luminance of all pixels in the image \cite{collewet2011photometricvs} addresses the issue of object limitations, but still requires camera calibration.  \cite{deguchi1996direct} on the other hand, represented images with principal component analysis that greatly reduces the dimensions and \cite{tahri2005pointvs} used a moments-based approach to extract features. All these methods still require fine-tuning for different applications.  
\begin{figure*}
\centering
\includegraphics[width=.8\textwidth]{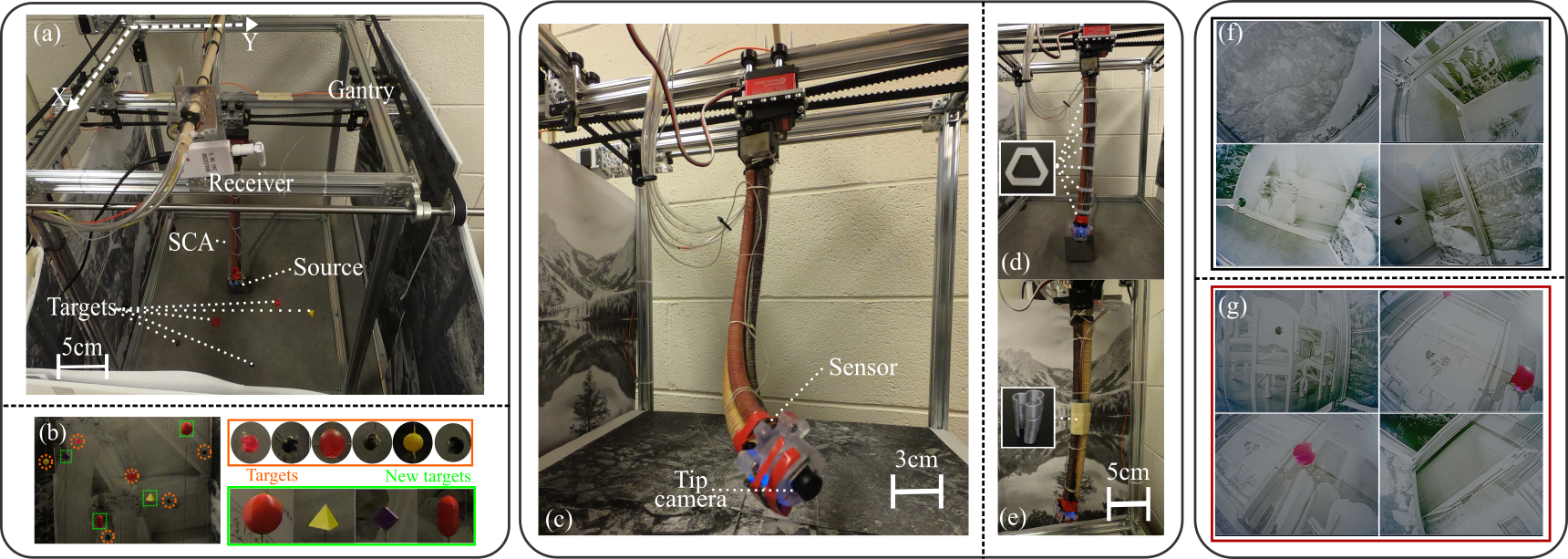}
\caption{Experimental setup: (a) BR$^2$ SCA attached to a rotating servo that can move in X and Y direction in the gantry along with the targets and the wireless receiver to receive the tip camera image. (b) Four new targets (not seen in training) along with the targets used for training. (c) BR$^2$ SCA  with the camera attached to the tip using a 3D printed casing. (d) SCA with uniform loads distributed along its length (inset: silicone cast ring weighing 1.4 grams). (e) SCA with the central region constrained with a rigid 3D printed part. (f) Four sample images used for the training with first background and (g) four sample images used for the training with the second background.}
\label{fig:exp_setup}
\end{figure*}

As feature extraction techniques in computer vision improved with the advent of neural networks, so did their applicability in visual servoing. A related paper in this area \cite{quentin2018deepnnvs} made use of deep neural networks like AlexNet \cite{krizhevsky2012imagenet} and VGG \cite{simonyan2014very} to learn the relative pose that is fed into the control policy. Our work is primarily inspired by this approach. More advanced deep learning models like LSTMs \cite{hochreiter1997lstm}, GANs \cite{goodfellow2014generative} are seen in \cite{sadeghi2017sim2real}, \cite{pedersen2020graspingunknownvs} respectively. \cite{paradis2020intermittent} on the other hand implemented a hybrid control policy where open-loop odometry was used as a coarse policy and a visual feedback policy was used to close the final error gaps to reach the targets. However, the above-mentioned works focused mainly on rigid arm visual servoing for which the system model is already known. 

Visual servoing for SCAs has gained a lot of traction recently, due to their difficulty in modeling and pose control. Works like \cite{fan2019underwatervs}, \cite{fan2021cabledriven} used a fixed camera (eye-to-hand) to capture the pose and curvature of the soft-arm to perform image-based visual servoing. Additional sensor assistance-based visual servoing was performed in \cite{wang2020} in order to track the camera motion but was limited to 2D space. In this letter, we are interested in eye-in-hand image-based visual servoing in a 3D framework. We rely on neural-networks to handle the feature extraction and mapping to actuation. The control policy then computes the error between the predicted actuations of current and target images. 

 \subsection{Overview}
We propose two approaches, \textit{integrated} and \textit{modular}, to estimate the pose of the soft manipulator, and control it using visual servoing in a structured environment. The integrated approach predicts the actuation directly for a given input image, which is useful when the environment is changed. The modular approach on the other hand, first predicts the pose for the given image and then maps the predicted pose to actuation which is particularly useful when the SCA is changed. Both these frameworks take a single RGB image, \textit{I}, and predict the control inputs (actuations) required to reach the corresponding pose of the soft arm (current pose). Using this information we calculate the error in the geometrical features of the current and target images, as well as the error between the current and target actuations. These errors are reduced by using visual feedback to estimate the control commands needed to reach the desired target pose. Through experiments, we show that both the approaches perform well, with the \textit{integrated} approach being robust to various changes such as light intensity, diminution of SCA, added weights, etc. 
Fig. \ref{fig:Full system}(a) shows the overall workflow of the proposed approach. 

% \begin{figure*}[t]
% \centering
% 		\hspace*{-.55cm}\centering\includegraphics[width=2.14\columnwidth]{Figures/Full_system_final.png}
% 		\caption{Visual Servoing Workflow}
% 		\label{fig:workflow}
% \end{figure*}
\section{Methods}
\subsection{Experimental setup}

The experimental setup consists of five connected systems: Soft Continuum Arm (SCA), gantry, electrical control board, computers, and magnetic sensor. The SCA (Fig. \ref{fig:exp_setup}(c)) is made of three Fiber Reinforced Elastomeric Enclosures (FREE)\cite{uppalapati2018towards} - one bending, two rotational (one clockwise(CW) and another counterclockwise (CCW)) and is referred to as a BR$^2$ \cite{uppalapati2021}. It has an individually controllable pneumatic actuator for each FREE.  The gantry (Fig. \ref{fig:exp_setup}(a)) adds three degrees of freedom (DOF) to the SCA via an $X$ and $Y$ rail and a rotational mount ($\theta$) for the SCA.  The $X$ and $Y$ rails are belt driven by stepper motors (NEMA 17) and have an $X$ travel of 45 cm and a $Y$ of 42 cm with the origin defined by limit switches. Positioning on the gantry is open loop and was reset between tests and data collection runs to reduce error accumulation. A servo motor (DS3218MG, DSSERVO) joins the SCA to the gantry and controls $\theta$($\pm$90\degree). Together the SCA and gantry provide five DOF: bending, rotation, theta, $x$ and $y$ translation.  Note that rotation is treated as one DOF as the two rotating FREEs are never actuated simultaneously.  The CW and CCW rotations are distinguished by positive or negative value. 
    
The electrical control board contains a pressure regulator (ITV0031-2UBL, SMC) for each FREE in the SCA, a PWM control board (PCA9685, Adafruit) for the servo and two stepper drivers (Big Easy Driver, SparkFun) to control the gantry translation.  These devices are operated by a Raspberry Pi 4 (8GB) and an Intel NUC (NUC7i7), both running Ubuntu 18.04 with ROS Melodic.  The Raspberry Pi is used to interface with the electrical control board while the NUC is used for the computationally intense control loop.  The two computers communicate via ROS multimaster. A magnetic sensor (micro sensor 1.8, Patriot SEU, Polhemus), attached to the SCA, provides pose information about the tip of the SCA relative to a fixed source (TX1, Polhemus) origin that is placed at the center of gantry base.
    
\subsection{Data collection}
We  mounted  a 1200  TVL  camera (Caddx Firefly, Micro FPV Camera w/ VTX), which is a low-cost, lightweight (4.2 grams), small form-factor camera on the distal tip of the SCA and collected images from the camera at various views by moving the soft arm and gantry. The setup of the soft arm is given in Fig. \ref{fig:exp_setup}(c). The process is automated and the inputs are given in the form of actuations, such as pressures ($b$, $r$), $x$, $y$ and angle (theta). Images of  the  scene  are  captured  at discrete configurations throughout the workspace while state data (actuations and sensor readings) is collected to self-annotate the images. A few examples of images taken by the camera are shown in Fig. \ref{fig:exp_setup} (f) and (g). The images have a resolution of 640x480 pixels.

\begin{figure*}
    \centering
    \includegraphics[width=.85\textwidth]{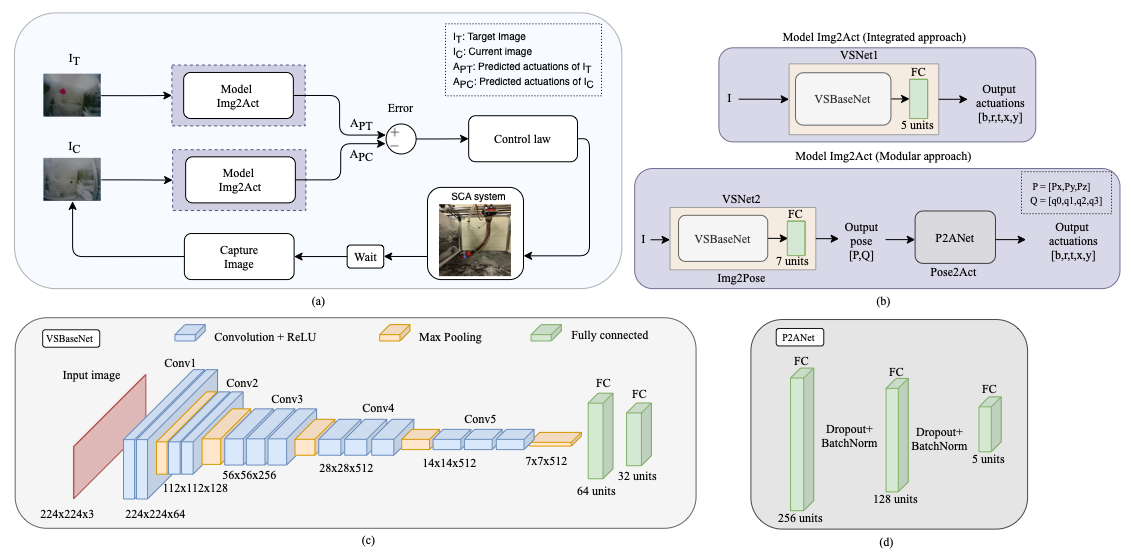}
    \caption{(a) Workflow of our method to reach the target image given current image. (b) Two different approaches (modular and integrated) for obtaining a mapping from image to actuations (Img2Act) (c) Network architecture of VSBaseNet and (d) Network architecture used for the pose to actuation mapping (P2ANet).  }
    \label{fig:Full system}
\end{figure*}

\subsection{Network Architecture}
Due to the ability of Deep Convolutional Neural Networks (CNNs) to automatically extract features from large training datasets, they have shown to be effective in various computer vision applications, such as image recognition  \cite{krizhevsky2012imagenet}, segmentation and also have been studied to estimate the pose of a robot manipulator given image inputs \cite{quentin2018deepnnvs}. Inspired by this, we use VGG16 \cite{simonyan2014very}, to estimate the input actuation values required to reach a specific pose of the soft manipulator arm using image inputs. VGG16 \cite{simonyan2014very} is originally trained for classification task on 1.2 million ImageNet images that has around 138 million parameters. Since our task is not exactly similar to the image classification, we modified the final few layers, performed transfer learning by using previously trained VGG16 weights on some layers and fine-tune it on our data which effectively helped the network to learn new features pertaining to our task. We also found that freezing the first 12 layers of the network and retraining the remaining layers gave optimal results in terms of loss and error. In addition to this, we added 2 fully connected layers (with 64, 32 units, respectively) with ReLU non-linearity. To aid regularization, we added batch normalization layers, dropout layers after the dense layers and also applied $l_1$ and $l_2$ regularizers to all the dense layers to decrease overfitting with $0.0001$ and $0.0005$ as their respective regularization factors. We call this network VSBaseNet. Fig. \ref{fig:Full system}(c) shows the complete network architecture of the base network, VSBaseNet. 

Two different approaches namely, \textit{integrated approach} and \textit{modular approach}, were used to predict the actuations from the input image. These two approaches were implemented and tested in order to see their effectiveness in various scenarios as shown in section \ref{results}. The workflows of both the approaches are given in Fig. \ref{fig:Full system}(b) and their network architectures details are given below. 

\noindent\textbf{Integrated Approach:}
In the integrated approach, the network directly outputs the actuations given an input image, $I$. Here, the network used is VSNet1 which consists of the base network, VSBaseNet, along with a dense output layer with sigmoid activation. Since we were dealing with a regression task, the final dense layer consists of five units that output 5 floats corresponding to the five input actuations: bending ($b$), rotation ($r$), theta ($t$), and the gantry ($x$ and $y$). The details of VSNet1 are given in Fig.  \ref{fig:Full system}(b) and (c).   
 
\noindent\textbf{Modular Approach:}
For the modular approach, we divided the image-to-actuation step in two parts (modules): image-to-pose, and pose-to-actuation. The image-to-pose (Img2Pose) module takes in a single image (taken at the current arm pose), $I$, and outputs the pose information. The VSNet2 network is used to take an input image and output the pose in the form of a vector comprised of the position and orientation (quaternion) information, $[p_x, p_y, p_z,q_0, q_1, q_2, q_3]$. This pose information is fed as input to the P2ANet network which predicts the corresponding mapping of actuation inputs in the form of another vector consisting of actuation values $[b, r, t, x, y]$. The network architecture of VSNet2 is similar to VSNet1 (it uses the same base network, VSBaseNet), except the last (output) layer, which has 7 units corresponding to the 7 output floats. The P2ANet consists of 3 dense layers with 256, 128, and 5 units respectively along with ReLU non-linearity in the first dense layer and sigmoid activation the last layer. We also added batch normalization and dropout layers to aid regularization. The network architecture of VSNet2 and P2ANet is given in Fig.  \ref{fig:Full system}(b), (c) and (d). 

\subsection{Training}

\subsubsection{Dataset}
Using our self-annotated data collection method, a total of 7980 images corresponding to different poses were collected. The absolute pose data with respect to the initial configuration was also noted for each of the images. We used electromagnetic tracking (Patriot SEU, Polhemus) with a short-range source (TX1, tracking area 2 to 60 cm) to get the ground truth absolute pose. This sensor is flexible, lightweight ($<$ 2 g), has a positional accuracy of less than 1mm and does not hinder or alter the performance of the soft arm. The signal from the sensor provides the real-time spatial coordinates of the soft arm end in the form of $[x,y,z,quaternion]$, while $[theta,r_1,r_2,b]$ come from the requested actuations.

In our approaches, we used image data to predict the actuations (integrated approach) or pose (modular approach) of the soft arm. The range of values for each of the 5 actuations were as follows: Bending ($b$): 14 to 22 psi (discrete values with steps of 2 psi) (96.5 to 151.7 kPa in 13.8 kPa steps); Rotation ($r$): -18 to 18 psi (discrete values with steps of 2 psi)(-124.1 to 124.1 kPa in 13.8 kPa steps); Theta ($t$): +6 to -6 degrees (discrete values with steps of 2); $x$: 14, 16 and 18 cm (discrete values); $y$: 14, 16, 18, and 20 cm (discrete values).

The dataset is divided into training, validation and testing sets with 4910, 1676, and 2394 images respectively. The ground truth values for the integrated approach consist of the absolute actuation values corresponding to the pose of the soft arm for each image. A CSV file containing 5 columns corresponding to each of the actuation values was created, and then split into training, validation and testing label files for training purposes. This method was repeated for the image-to-pose part of the modular approach where the ground truth values consisted of the pose information. This entire data collection process is automated.  

\subsection{Loss Function and Optimization}

Our network takes in a single image (taken at the current arm pose), $I$, and outputs the absolute actuation values required to reach that pose. Since this is a regression problem, the last layer of the network outputs floats. The output of the network is in the form of a vector comprising of either the pose $(p_x, p_y, p_z, q_0, q_1, q_2, q_3)$ or the 5 actuations $(b, r, t, x, y)$. To regress absolute values of pose or actuations, we use the mean-squared error (MSE) loss function which computes the mean of squared errors between the ground truth values and the predictions.

\vspace{-0.2cm}
\begin{equation}
    loss(I) = \frac{1}{n}\sum_{i=1}^{n}(Y_i - \hat{Y_i})^2 
\end{equation}

Here, $Y_i$ corresponds to the ground truth actuations whereas, $\hat{Y_i}$ corresponds to the predicted actuations of the input images. We experimented with SGD and Adam optimizer for training and found that Adam optimizer converged faster and with less oscillation. We achieved best results using a time based learning rate scheduler with an initial learning rate of 0.01 and number of epochs as 150. The learning rate at each epoch was calculated as:
\vspace{-0.2cm}
\begin{equation}
    \eta_{n} = \eta_{n-1} * \frac{1}{1+decay*n}
\end{equation}
where $\eta_{n-1}$ is the learning rate of the previous epoch, and $n$ is the current epoch number. The value of decay is normally implemented as:
\begin{equation}
    decay = \frac{\eta_0}{N}
\end{equation}
where $\eta_{0}$ is the initial learning rate and $N$ is the total number of epochs. We trained the model for 150 epochs after saturation is reached. We used a batch size of 128 to help generalizing the model better. Using a lower or a higher batch size caused the validation loss to fluctuate. 

\begin{figure*}[t]
    \centering
    \includegraphics[width=.8\textwidth]{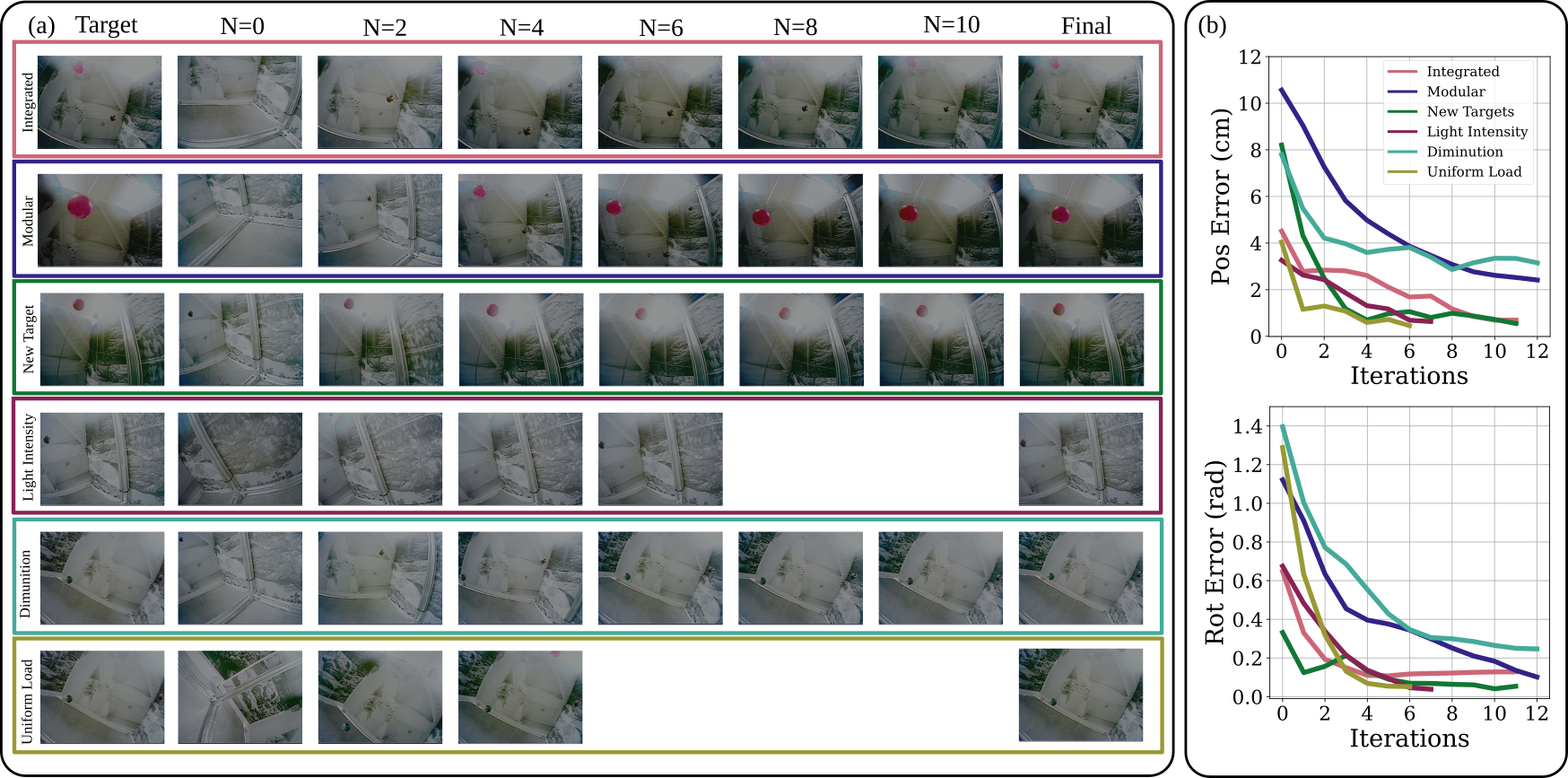}
    \caption{Results (for one case each): (a) The target, current images at different iterations (denoted by N) and the final image when the stopping condition $MSE_a < 5$ was reached and (b) the corresponding position and rotation error over iterations for integrated, modular, new targets, light intensity, diminution and uniform load.} 
    \label{fig:results}
\end{figure*}
\subsection{Control Architecture}
% Due to the inherent non linearity of the SCA, effect of internal and external disturbances like weights the SCA may not necessarily reach the same pose when the training data is collected. In addition, there will also be some errors due to the trained model. To overcome these unpredictable characteristics of the SCA behaviour we use a control law to make the tip of the SCA to converge to the target image ($I_T$). The following control law is used as shown in the Fig.\ref{fig:Full system}(a):
% \begin{align}
%     A_{RC}(k+1) = A_{RC}(k)-\lambda(A_{PC}(k)-A_{PT})
% \end{align}
% where $A_{RC}(k), A_{PC}(k) and A_{PT}$ are the current actuation's to the soft arm, predicted actuation's for the current image and predicted actuation's for the target image at step $k$. As the error between the predicted actuation's for the current image and target image reduces to zero the SCA reaches a steady state and also implies the tip is seeing the target image. $\lambda$ is the proportional gain ($> 0$) used for efficient convergence. 

There are two possible sources for open loop errors in the system, (i) Non repeatability due to hysteresis could lead to a different end effector position for the same input actuations, which could also be dependent on the path taken by the manipulator\cite{uppalapati2021}. (ii)  Inaccuracies in the trained model to fit the pose to actuations could also lead to large deviations from the target. To overcome these inherent errors, we integrated the following control update, where the error between the current predicted actuations and the target actuations at various iterative steps are fed back into the input until the tip converges to the target image ($I_T$) within reasonable accuracy, as shown in the Fig.\ref{fig:Full system}(a):
\begin{align}
    A_{RC}(k+1) = A_{RC}(k)-\lambda(A_{PC}(k)-A_{PT})
\end{align}
where $A_{RC}(k), A_{PC}(k)$ and $A_{PT}$ are the current actuations to the soft arm, predicted actuations for the current image and predicted actuations for the target image at step $k$.It must be noted that the arm operates in a  quasi-static manner in each iteration step and at the end of each step $k$, it is made to reach static equilibrium where all the external forces are balanced by the actuation forces. The current image for the next iteration is taken only after this equilibrium is reached after 6 seconds and hence the \textit{wait} after system actuations as shown in figure \ref{fig:Full system}(a). As the error between the predicted actuations for the current image and target image reduces to zero, the SCA tip reaches its target position (or the tip camera views the target image). $\lambda$ is the proportional gain ($> 0$) used for efficient convergence. The overall gain $\lambda$ used is decoupled to two different gains, $\lambda_r$ for the $x,y$ and $\theta$ variable and $\lambda_s$ for the $b,r$ variables in order for efficient and smooth convergence.

\section{Results}\label{results}
In this section, we describe the different scenarios used to validate the approaches detailed in Section II on the SCA. 

\subsection{Formulation of normalized, unitless MSE$_a$ metric}

A normalized MSE metric that is scale-invariant and unit-less is formulated in order to represent the accuracy of the system is shown in Eq. \ref{eq:mse}. In this equation, each term is divided by the resolution i.e., the minimum change a state can undergo. Here, N = 5 corresponds to the 5 actuations - b (kPa) (i=1), r (kPa) (i=2), t (radians) (i=3), x (m) (i=4), y (m) (i=5); $a_{observed}$ is observed actuation, $a_{target}$ is target actuation and $a^i_k$ = 0.1 is the scaling factor $\forall i \in \{1, 2, 3, 4,5\}$. All states are rounded off to their first decimal point and hence 0.1 (0.1 kPa, 0.1, radians, 0.1 m) is the scaling used. Based on this metric, we define the stopping condition for all the tests conducted to be $MSE_a < 5$ or when the number of iterations (N) reaches 15. These values were empirically decided with two criteria: a) reduce the translation and rotation error and b) reach the target image in a reasonable number of iterations.

\begin{equation} \label{eq:mse}
    MSE_a = \frac{1}{N}\sum_{i=1}^{5}\left(\frac{a^i_{observed} - a^i_{target}}{a^i_k}\right)^2
\end{equation}

\subsection{Estimation of $\lambda_s$ and $\lambda_r$}
The different actuations have a disproportionate effect on the SCA tip position. For example, a small change in $x$ or $y$ position will have a larger effect on the SCA tip than a similar change of the pressure in the SCA. The tip position is also dependent on the current shape of the SCA. It is empirically obtained that the number of iterations required to reach a test image to obtain the actuation error ($MSE_a$)  less than 5 is faster for values of $\lambda_r$ and $\lambda_s$ in the range of [0.5, 0.7] and [0.6, 0.8]. Based on this test case, the values of $\lambda$ for all the following validation tests is set to $[\lambda_r,\lambda_s] = [0.6, 0.7]$.

% \begin{figure}[!h]
%     \centering
%     \includegraphics[width=.48\textwidth]{Figures/lambda_big_times.png}
%     \caption{Total number of iterations for converging to the target image (stopping condition $MSE_a < 0.1$) for  values of $\lambda_s$ and $\lambda_r$ varying from 0.2 to 1 (in steps of 0.2) }
%     \label{fig:lambda}
% \end{figure}

\subsection{Integrated approach}
Thirty (n = 30) random points in the operating range/workspace of the SCA system were collected and their pose ($x,y,z, q0,q1,q2,q3$) information is recorded with the Polhemus magnetic sensor. VSNet1 (shown in Fig.\ref{fig:Full system}(b)) is used for reaching the desired target images. For each test, the SCA system starts with a random initial configuration. 

The target image, current images at different iterations, and the final image (when the stopping condition of $MSE_{a} < 5$ was reached) for one of the test cases is shown in Fig.\ref{fig:results}(a). It took eleven iterations for it to reach the desired stopping condition. From the position and rotation error plots in Fig.\ref{fig:results}(b), it can be observed that the error was reduced to less than 2 cm in six iterations. In the remaining iterations, the system transitions to further reduce the error. The accuracy of this approach is also shown with the quantitative metrics of average MSE in actuations, average Euclidean distance error, and average rotation error between the final and target image for all the 30 tests as reported in Table \ref{tab:approach_comp}. We would like to highlight that for two of the test cases where the arm looks at the ground with no features initially, it reached with $77$ and $30.7$ $MSE_a$ at the 15th iteration leading to average $MSE_a = 5.587 > 5$.

\begin{figure*}[t]
    \centering
    \includegraphics[width=.88\textwidth]{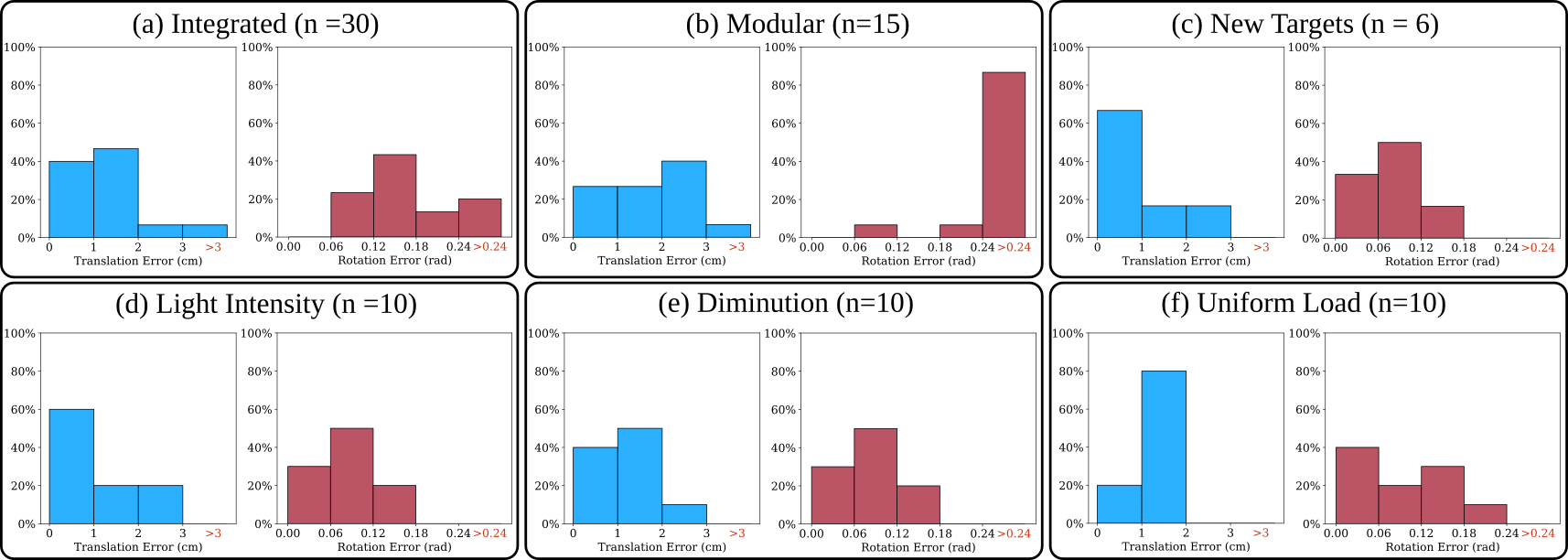}
    \caption{Histogram of translation and rotation errors obtained for the test cases of (a) Integrated (30 points), (b) Modular (15 points), (c) New Targets (6 points), (d) Change in light intensity (10 points), (e) Diminution of SCA functionality (10 points), and (f) Uniform load (n = 10 points).}
    \label{fig:hist_results}
\end{figure*}

Figure \ref{fig:hist_results}(a) shows the histogram of translation and rotation errors for the 30 test points. Translation error is calculated using the Euclidean distance between the ground truth $(p_x, p_y, p_z)$ position (obtained from the Polhemus magnetic sensor) of the target image and final image for each test. Rotation error on the other hand is obtained using Euler's Axis-angle representation where $R_1, R_2$ are rotation matrices at the target and final images respectively.  The quaternion pose information obtained by the Polhemus sensor is first converted to rotation matrix  in order to use the Eq.\ref{eq:rot_error}.

\begin{equation}\label{eq:rot_error}
e(R_1, R_2) = cos^{-1}\Big(\dfrac{trace(R_1 R_2^T) - 1}{2}\Big)
\end{equation}

% \begin{equation}\label{eq:quat2R}
% R = \begin{bmatrix}
% 1 - 2(q_2^2 + (q_3^2) & 2(q_1 q_2 - q_3 q_0) &  2(q_1 q_3 + q_2 q_0) \\
% 2(q_1q_2 + q_3 q_0) & 1 - 2(q_1^2 + q_3^2) & 2(q_2 q_3 - q_1 q_0)\\
% 2(q_1q_2 - q_3q_0) & 2(q_2 q_3 + q_1 q_0) & 1 - 2(q_1^2 + q_2^2)
% \end{bmatrix}
% \end{equation}

\begin{table}
  \caption{Comparison between Integrated and Modular approach}
  \label{tab:approach_comp}
  \setlength{\tabcolsep}{4.2pt}
  \begin{tabular}{lccc}
    \toprule
    \multicolumn{1}{p{0.3\columnwidth}}{\centering \hspace{-0.1cm}Method and number of tests (n)}
    & \multicolumn{1}{p{0.165\columnwidth}}{\centering Avg. MSE$_a$ (normalized - no units)}
    & \multicolumn{1}{p{0.165\columnwidth}}{\centering Avg. Euclidean dist. error (cm)}
    & \multicolumn{1}{p{0.165\columnwidth}}{\centering Avg. rotation error (radians)}\\
    
    \midrule
    Integrated approach (n=30) & 5.587* & 1.6481 & 0.2325 \\
    Modular approach (n=15) & 6.489* & 1.8002 & 0.4261 \\
  \bottomrule
\end{tabular}
\end{table}

% \begin{table}
%   \caption{Comparison between Integrated and Modular approach}
%   \label{tab:approach_comp}
%   \setlength{\tabcolsep}{4.2pt}
%   \begin{tabular}{lccc}
%     \toprule
%     \multicolumn{1}{p{0.3\columnwidth}}{\centering \hspace{-0.1cm}Method and number of tests (n)}
%     & \multicolumn{1}{p{0.165\columnwidth}}{\centering Avg. MSE$_a$ (normalized - no units)}
%     & \multicolumn{1}{p{0.165\columnwidth}}{\centering Avg. Euclidean dist. error (cm)}
%     & \multicolumn{1}{p{0.165\columnwidth}}{\centering Avg. rotation error (radians)}\\
    
%     \midrule
%     Integrated approach (n=30) & 5.587 & 1.6481 & 0.2325 \\
%     Modular approach (n=15) & 6.489 & 1.8002 & 0.4261 \\
%   \bottomrule
% \end{tabular}
% \end{table}

\begin{table}
  \caption{Results of experiments (integrated approach)}
  \label{tab:results_exp}
  \setlength{\tabcolsep}{2.2pt}
  \begin{tabular}{lccc}
    \toprule
    % \multicolumn{1}{p{1.5cm}}{\centering Method}
    % \multicolumn{1}{p{1.5cm}}{\centering Method}
    \multicolumn{1}{p{3cm}}{\centering  \hspace{-0.1cm}Method and number of tests (n)}
    & \multicolumn{1}{p{1.36cm}}{\centering Avg. MSE$_a$\\(normalized - no units)}
    & \multicolumn{1}{p{1.36cm}}{\centering Avg. Euclidean dist. error (cm)} 
    & \multicolumn{1}{p{1.36cm}}{\centering Avg. rotation error (radians)}\\
    \midrule
    New targets in workspace (n = 6) & 2.828 & 1.1108 & 0.0858 \\
    Lighting changes (n = 10) & 3.485 & 1.0690 & 0.0857 \\ %no change
    Diminution (n = 10) & 3.777 & 1.4491 & 0.1350 \\
    Uniform load (n = 10) & 3.296 & 1.3274 & 0.0975 \\
    Change in background (n = 5) & 0.778 & 1.4212 & 0.1252 \\
  \bottomrule
\end{tabular}
\end{table}

% \begin{table}
%   \caption{Results of experiments (integrated approach)}
%   \label{tab:results_exp}
%   \setlength{\tabcolsep}{2.2pt}
%   \begin{tabular}{lccc}
%     \toprule
%     % \multicolumn{1}{p{1.5cm}}{\centering Method}
%     % \multicolumn{1}{p{1.5cm}}{\centering Method}
%     \multicolumn{1}{p{3cm}}{\centering  \hspace{-0.1cm}Method and number of tests (n)}
%     & \multicolumn{1}{p{1.36cm}}{\centering Avg. MSE$_a$\\(normalized - no units)}
%     & \multicolumn{1}{p{1.36cm}}{\centering Avg. Euclidean dist. error (cm)}
%     & \multicolumn{1}{p{1.36cm}}{\centering Avg. rotation error (radians)}\\
%     \midrule
%     New targets in workspace (n = 6) & 4.628 & 1.1108 & 0.0858 \\
%     Lighting changes (n = 10) & 3.485 & 1.0690 & 0.0857 \\ %no change
%     Uniform load (n = 10) & 3.296 & 1.3274 & 0.0975 \\
%     Diminution (n = 10) & 3.777 & 1.4491 & 0.1350 \\
%     Change in background (n = 5) & 0.778 & 1.4212 & 0.1252 \\
%   \bottomrule
% \end{tabular}
% \end{table}

\subsection{Modular approach}
The modular approach (as in Fig.\ref{fig:Full system}(b)) was tested on fifteen random points (n = 15) in the workspace within the range of the SCA and the gantry. The pose information for all the test images was recorded using the Polhemus magnetic sensor. VSNet2 predicts the pose given an input image and the P2ANet outputs the corresponding actuations for the predicted pose. The quantitative metrics using the 15 tests is given in Table \ref{tab:approach_comp}. The target image, current images at different iterations, and the final image (when the stopping condition of $MSE_{a} < 5$ was reached) for one of the test cases is shown in Fig.\ref{fig:results}(a). As seen in Fig.\ref{fig:results}(a), the final image obtained after converging in 12 iterations is a little farther from the desired target image, however the orientation is much closer to the desired orientation using this method. We also observed that two of the tests with $MSE_a$ of $172.3$ and $44.7$ at the 15th iteration, resulting in average $MSE_a = 6.489 > 5$. Fig.\ref{fig:hist_results}(b) shows the translation and rotation errors for the 15 test points.  

\subsection{New targets}
The integrated approach is tested new targets (as shown in Fig. \ref{fig:exp_setup}(b)) inserted in the workspace. Six target images (n = 6) were randomly collected, out of which three images contained the new target alone, and remaining three images contained both new and old targets (included during training). The target image, current images at different iterations, and the final image (when the stopping condition of $MSE_{a} < 5$ was reached) for one of the test cases is shown in Fig.\ref{fig:results}(a). As seen in the position error plot in Fig. \ref{fig:results}(b), the error reduced to less than 2 cm in three iterations and converges to the new target image in 11 iterations. The quantitative metrics using the six tests are given in Table \ref{tab:results_exp}. Fig. \ref{fig:hist_results}(c) shows the histogram of translation and rotation errors for the six test points.

\subsection{Robustness to light changes}
The robustness of our integrated approach against light exposure changes was tested with an extra light source in the environment, thus making it brighter. Tests were conducted at an average illuminance of 341.4 lx compared to 155.4 lx for training and other testing (Light Meter Model R8130, Reed Instruments). The results for one case are shown in Fig. \ref{fig:results}(a)-(b). For this case the target image was reached in six iterations. The quantitative metrics using the ten tests are given in Table \ref{tab:results_exp}. Fig. \ref{fig:hist_results}(d) shows the histogram of translation and rotation errors for the ten test points (n = 10).

\subsection{Effect of diminution}
For this experiment, we restricted the functionality of the SCA by attaching 3D printed clips to its mid section as shown in Fig. \ref{fig:exp_setup}(e). These clips restrict the bending functionality of the SCA in the sealed section of the arm. The integrated approach was tested on 10 different random images. The results of one test case are shown in Fig. \ref{fig:results}(a) and (b). As seen in the Fig. \ref{fig:results}(b), the SCA reached the target image in 12 iterations.  The quantitative metrics using the 10 tests are given in Table \ref{tab:results_exp}  along with the histogram of translation and rotation errors for the 10 test points (n = 10) in Fig. \ref{fig:hist_results}(e).

\subsection{Uniform load}
Six uniform rings of 1.4 grams each were added on to the SCA equidistantly along the length as shown in Fig.\ref{fig:exp_setup}(d). The rings were fabricated with silicon and thus owing to flexibility of silicon, these rings don't affect the functionality of the SCA at the added locations. A total of ten experiments were conducted. The integrated approach was used for this experiment, where results of one of the tests with stopping condition $MSE_{a} < 5$ is shown in Fig. \ref{fig:results}(a). The target was reached accurately with loads in six iterations. The total added weight is around 25\% of the total weight of the SCA.  The quantitative metrics using the ten tests (n = 10) are given in Table \ref{tab:approach_comp} along with the histogram of translation and rotation errors for the ten test points in Fig.   \ref{fig:hist_results}(f).
% \begin{figure}[htp!]
% \centering
% 		\hspace*{-.55cm}\centering\includegraphics[scale = 0.35]{Figures/modular_workflow.png}
% 		\caption{Modular approach workflow}
% 		\label{fig:mod_workflow}
% \end{figure}

\section{Adaptability to a new environment}
\begin{figure}[htbp]
    \centering
    \includegraphics[width=2.26 in]{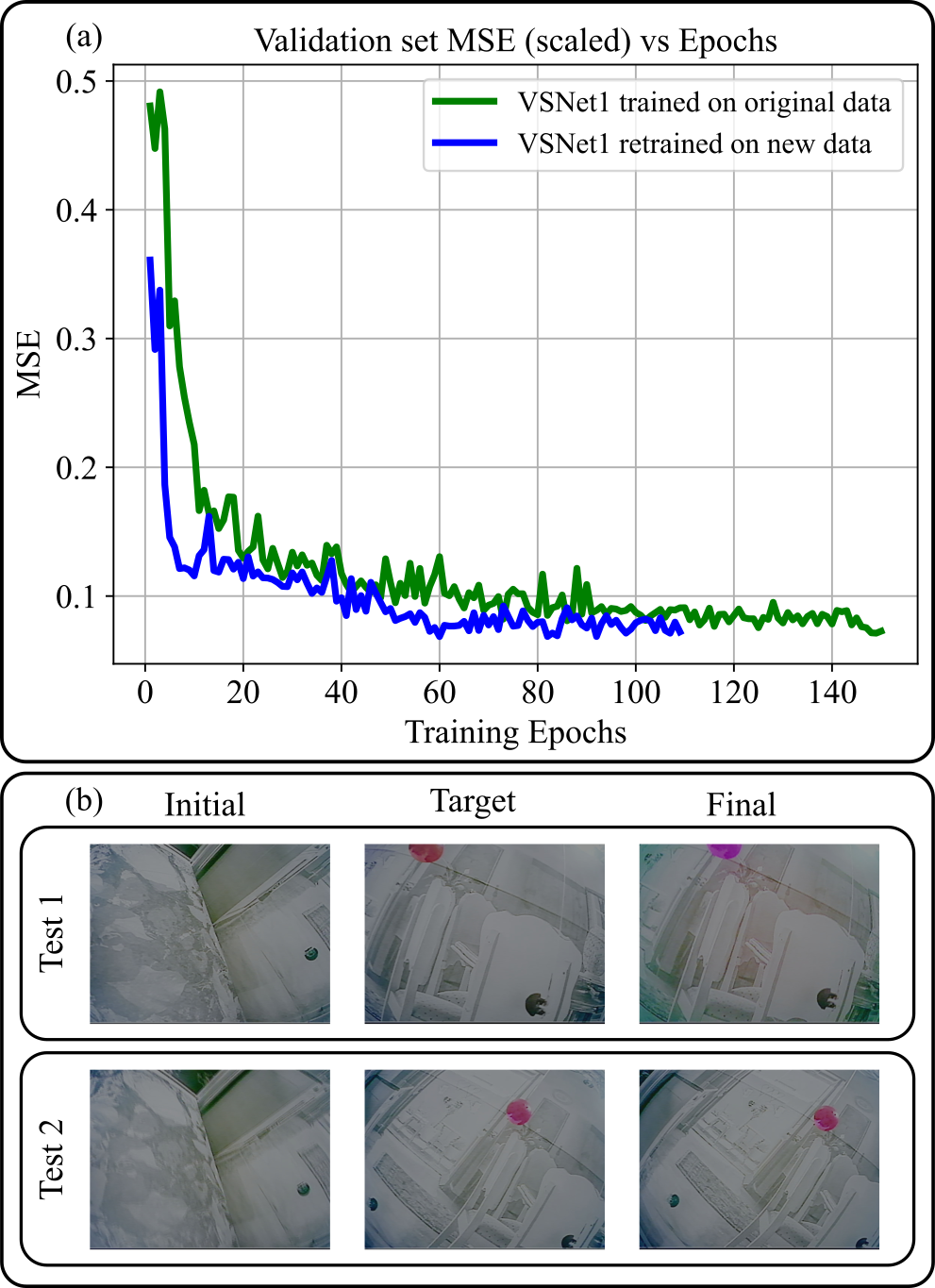}
    \caption{Results: (a) Validation set MSE trend for original data trained on VSNet1, and new data retrained on VSNet1 and (b) The initial, target and the final image when the stopping condition $MSE_a < 1$ was reached.}
    \label{fig:val_mse}
\end{figure}

In order to test the transferability and adaptability of the system to new environments, we changed the background of our structured environment. We added previously unseen images in the background of our setup and additionally included images on the ground (bottom of the environment).  With the new background, data was recollected as described in Section IID. Our model was retrained on the new background data, with weights initialized as the trained weights from the original VSNet1. 
Five experiments were conducted using the retrained model in the new environment, keeping the stopping condition as $MSE_a < 1$ and maximum iterations as $15$. The results of two cases are shown in Fig. \ref{fig:val_mse} (b) which took 27 and 23 iterations respectively, to reach the stopping condition. The average number of iterations to reach the stopping condition for all the tests was 23. The mean translation error was 1.4212 cm and the mean rotation error was 0.1252 radians. We also observed that retraining VSNet1 took fewer steps and converged faster than before (converged in 110 epochs as opposed to 150 epochs from before). This can be seen from the validation set MSE graph in Fig. \ref{fig:val_mse}(a).

\section{Discussion}
In this paper, we demonstrate that visual servoing using deep neural networks leads to accurate and robust control of a soft continuum arm, which is otherwise known to be hard to control using model-based techniques. We showcased two approaches for deep-learning based visual servoing of SCAs, the first utilizing an  \textit{integrated} (image to actuation) approach, and the second utilizing a \textit{modular} approach (image to pose and pose to actuation). In the integrated approach as seen in Fig. \ref{fig:hist_results}(a), 90\% of the data has less than 2 cm translation error (approximately the diameter of the SCA) and 80\% less than 0.24 radians for the rotation. The test cases with higher error occurred on the extremities of the workspace (edge of the gantry in this case). Such errors are likely a result of no features in background in two different parts of the workspace causing the model to get confused between them. In these cases, the gantry bottom had a plain background and the model was confused for a similar image on the other corner of the gantry. This can be addressed by having a non-plain background on all sides of the operating region. Excluding these outliers reduces the average translation error to less than 1.4 cm.

 The modular approach can be useful when either the SCA is changed (by retraining P2ANet alone) or the background is changed (retraining VSNet2 alone). Although the modular approach does a reasonable job in reducing the errors for more than 50\% of the data, from Fig. \ref{fig:hist_results}(a) and (b) it was found to be less accurate compared to the integrated approach. This may be due to errors that accumulate due to the intermediate pose estimation step. We do note here that in both approaches our architecture directly computes the control actuation, as such, this indicates that deep learning based visual servoing can be directly utilized in a control architecture with a simple linear control law. The reasonable tracking from our architecture indicates that further optimization of control was not necessary for our problem setup which was focused on the static reach problem. However, optimization and learning-based-control could be interesting directions for future work in problems like dynamic tracking, or trying to reach objects that are not reachable with static actuation by using the arm's momentum.%, such as swinging the arm to build momentum 
%Although the modular approach does a reasonable job in reducing the errors for more than 50\% of the data, from Fig.\ref{fig:hist_results}(a) and (b) it can be observed that the integrated approach does better than the modular approach with less translation and rotation errors. The modular approach can be useful when either the SCA is changed (have to retrain only P2ANet) or the background is changed (have to retrain only VSNet2). This is a useful approach in cases where either of the above mentioned changes are of frequent occurrence.

From the histogram plots for different cases Fig. \ref{fig:hist_results}(c-f), the integrated approach is robust to several changes the SCA may encounter (such as loads, disturbances and diminution) for performing different real-world tasks. The approach is able to reach the target positions with errors less than 1.5 cm for more than 80\% of tests in all cases. In addition, unlike the previous work on the control of the BR$^2$ SCA \cite{satheeshbabu2020continuous}, the image based method also controls the orientation of the SCA where the rotation errors were less than $0.24$ radians for 100\% of the data and no abrupt changes in actuations were noticed leading to smooth convergence of the end effector to the target. Furthermore, the system worked satisfactorily well in a new environment, considering the model was not fine-tuned to the new dataset. The data collection was efficient for a new background since it's automated. We observed that retraining VSNet1 took fewer steps and converged faster. Since we had retrained the model with images where the ground is visible, the system was able to converge upon encountering the ground during testing. We performed experiments on the new background with a more rigid stopping condition ($MSE_a < 1$) and found that our method is capable of performing more accurately with a stricter stopping condition. We also tested a few points in the new background with the previous model (trained on the original dataset), but it did not converge. This ascertains that retraining the VSNet1 with new data was required. Since we have a self-supervised system, collecting data and retraining on a new background can be done in a few hours.        

\section{CONCLUSION}
To conclude, we demonstrated that visual servoing with deep learning-based architectures leads to a reliable reach-control of soft continuum arms, which are otherwise known to be difficult to control. Our method includes a feedback controller, on top of our modified VGG16-based image-to-actuation predicting model, to accommodate for hysteresis present in the soft-arm as well as the inaccuracies in the actuation predictions. We demonstrated our method in static reach problems in structured non-changing environments, which captures a large operational set for such arms.  In these environments, we showed the robustness of our approach through various types of experiments ranging from change in environment lighting, new targets in the environment, restricting the functionality of the arm to adding uniform load. Additionally, we not only control the position of the arm but also the orientation as compared to \cite{satheeshbabu2020continuous}. We also verified the transferablility of our neural network model to a new environment by changing the background images coupled with retraining. As a result, a huge advantage is that the users can easily re-purpose our system for various settings without any need for manual labeling since the data collection for training the prediction model is automated.

While we limited this investigation to the quasi-static response of the SCA, in the future we will explore visual servoing in dynamic environments for which we will leverage the recent advances in spatio-temporal neural networks \cite{hochreiter1997lstm}. In future work, we would like to validate the effectiveness of the modular approach by changing the SCA that has a different architecture than the $BR^2$ SCA used in this work. Furthermore, acquiring a target image is limited to random exploration or a teaching policy method currently. In future work, we would like to give a query object as the target to which the arm should reach \cite{sadeghi2019divis}. We also acknowledge that this work is restricted to controlling the soft arm moving with zero collisions with its environment. With obstacles, the data collection process will no longer be automatic as shown in this work. Therefore, in our future work, we intend to investigate visual servoing in cluttered environments where the soft arm leverages its flexibility and interaction with the obstacles in reaching desired regions.

%working environment is without obstacles. Therefore, we would also investigate visual servoing in a cluttered environment to reach desired regions while taking advantage of the flexibility of our soft-arm. 

\addtolength{\textheight}{-12cm}   % This command serves to balance the column lengths

%%%%%%%%%%%%%%%%%%%%%%%%%%%%%%%%%%%%%%%%%%%%%%%%%%%%%%%%%%%%%%%%%%%%%%%%%%%%%%%%

%%%%%%%%%%%%%%%%%%%%%%%%%%%%%%%%%%%%%%%%%%%%%%%%%%%%%%%%%%%%%%%%%%%%%%%%%%%%%%%%

%%%%%%%%%%%%%%%%%%%%%%%%%%%%%%%%%%%%%%%%%%%%%%%%%%%%%%%%%%%%%%%%%%%%%%%%%%%%%%%%

%%%%%%%%%%%%%%%%%%%%%%%%%%%%%%%%%%%%%%%%%%%%%%%%%%%%%%%%%%%%%%%%%%%%%%%%%%%%%%%%

\bibliographystyle{IEEEtran}{}
\bibliography{References}

\end{document}